\documentclass[letterpaper, 10pt, conference]{ieeeconf}      
 
\IEEEoverridecommandlockouts                              

\usepackage{algorithm}
\usepackage{multicol}
\usepackage{multirow}
\usepackage{algpseudocode}
\usepackage{color}
\usepackage{xcolor}
\usepackage{graphicx}
\usepackage{bm}
\usepackage{amsmath}
\usepackage{amssymb}
\usepackage{tabularx}
\usepackage{subcaption}
\usepackage{url}
\usepackage{cite}
\usepackage{prettyref}
\usepackage{booktabs}
\usepackage{siunitx}

\newrefformat{fig}{Figure~\ref{#1}}
\newrefformat{sec}{Section~\ref{#1}}
\newrefformat{tab}{Table~\ref{#1}}
\newrefformat{alg}{Algorithm~\ref{#1}}
\newrefformat{eqn}{Equation~\ref{#1}}

\usepackage[draft]{hyperref}
\hypersetup{
    colorlinks,
    linkcolor={blue!80!black},
    citecolor={blue!80!black},
    urlcolor={blue!100!black}
}

\makeatletter
\newcommand{\printfnsymbol}[1]{%
  \textsuperscript{\@fnsymbol{#1}}%
}
\makeatother

\title{\LARGE  \bf
Getting Robots Unfrozen and Unlost in Dense Pedestrian Crowds
} 

\author{Tingxiang Fan$^*$, Xinjing Cheng$^*$, Jia Pan$^\dagger$, Pinxin Long, Wenxi Liu, Ruigang Yang and Dinesh Manocha%
\thanks{$^*$ denotes equal contribution. T. Fan, X. Cheng, R. Yang are with the Robotics and Auto-Driving Lab, Baidu Research. J. Pan is with the Department of Mechanical and Biomedical Engineering, the City University of Hong Kong. P. Long is with Metoak Technology (Beijing) CO., LTD. W. Liu is with College of Mathematics and Computer Science, Fuzhou University. D. Manocha is with the University of Maryland, College Park. $^\dagger$ denotes the corresponding author. Email: jiapan@cityu.edu.hk }
}

\begin{document}
\maketitle

\begin{abstract}
We aim to enable a mobile robot to navigate through environments with dense crowds, e.g., shopping malls, canteens, train stations, or airport terminals. In these challenging environments, existing approaches suffer from two common problems: the robot may get frozen and cannot make any progress toward its goal, or it may get lost due to severe occlusions inside a crowd. 
Here we propose a navigation framework that handles the robot freezing and the navigation lost problems simultaneously. First, we enhance the robot's mobility and unfreeze the robot in the crowd using a reinforcement learning based local navigation policy developed in our previous work~\cite{long2017towards}, which naturally takes into account the coordination between the robot and the human. Secondly, the robot takes advantage of its excellent local mobility to recover from its localization failure. In particular, it dynamically chooses to approach a set of recovery positions with rich features. To the best of our knowledge, our method is the first approach that simultaneously solves the freezing problem and the navigation lost problem in dense crowds. We evaluate our method in both simulated and real-world environments and demonstrate that it outperforms the state-of-the-art approaches. Videos are available at \url{https://sites.google.com/view/rlslam}.

\end{abstract}



\section{Introduction}
\label{sec:intro}

Navigating a mobile robot in complex, cluttered, and dynamic environments has a wide variety of important applications. For instance, assistive robots working in malls, cafeterias, and hospitals can benefit from a robust navigation policy that allows for efficient and safe movement in unstructured environments with dense crowds. Such navigation algorithm is also desperately needed by social devices such as Alexia \footnote{\textbf{Alexa}: https://developer.amazon.com/alexa} and Jibo \footnote{\textbf{Jibo}: https://www.jibo.com/}. Due to the lack of mobility, they rely on far-field speech recognition and speech synthesis to communicate with users at a very low information rate. If being mounted on a mobile base with sophisticated navigation skills, they could move into the user’s close-proximity and interact with users via visual interfaces at a much higher information rate. In addition, the high mobility is beneficial for an automated warehouse, where a large number of robots need to coordinate with each other for efficient transportation. For accomplishing a high delivery throughput, every robot needs to continuously make progress toward its goal by passing through the cluttered and dynamic environment made by its fellow robots. 

Unfortunately, classical algorithms for navigation in dynamic environments suffer from two major impediments: the \textit{robot freezing} problem and the \textit{navigation lost} problem. The \textit{robot freezing} problem arises when the environment surpasses a certain level of dynamic complexity. Specifically, the motion planner decides that all forward paths are unsafe and thus forces the robot to come to a complete stop or to perform unnecessary maneuvers like oscillating between two directions to avoid collisions. Previous works attempt to unfreeze the robot in dense crowds by increasing the prediction accuracy of the moving agents, which unfortunately has shown to be insufficient due to the lack of coordination among agents~\cite{trautman2015robot,Trautman:2010:IROS,Trautman:2013:ICRA,luo2018porca}. This conclusion is also supported by the investigation of human behaviors in dense crowds~\cite{Helbing:1995:PRE,Helbing:2000:Nature}.
The \textit{navigation lost} problem arises when the robot fails to accurately localize itself in a given map due to the large localization uncertainty or error~\cite{Fox:1999:MCL}.
Most previous solutions to the navigation lost problem are passive methods. They assume that the robot motion and the pointing direction of the sensors cannot be controlled, and focus on selectively utilizing the sensor stream to minimize the localization uncertainty or error~\cite{Fox:1999:MLM,Tipaldi:2013:LLC,Sun:2016:IROS}. However, in highly dynamic scenarios with dense human crowds, the salient features necessary for localization may all be occluded and thus the robot must actively determine ``where to move'' to resolve occlusion and ``where to look'' to recover from the localization lost~\cite{Burgard:1997:AMR,Li:2016:ALD}.

\begin{figure*}
\centering
\includegraphics[width=1.0\linewidth]{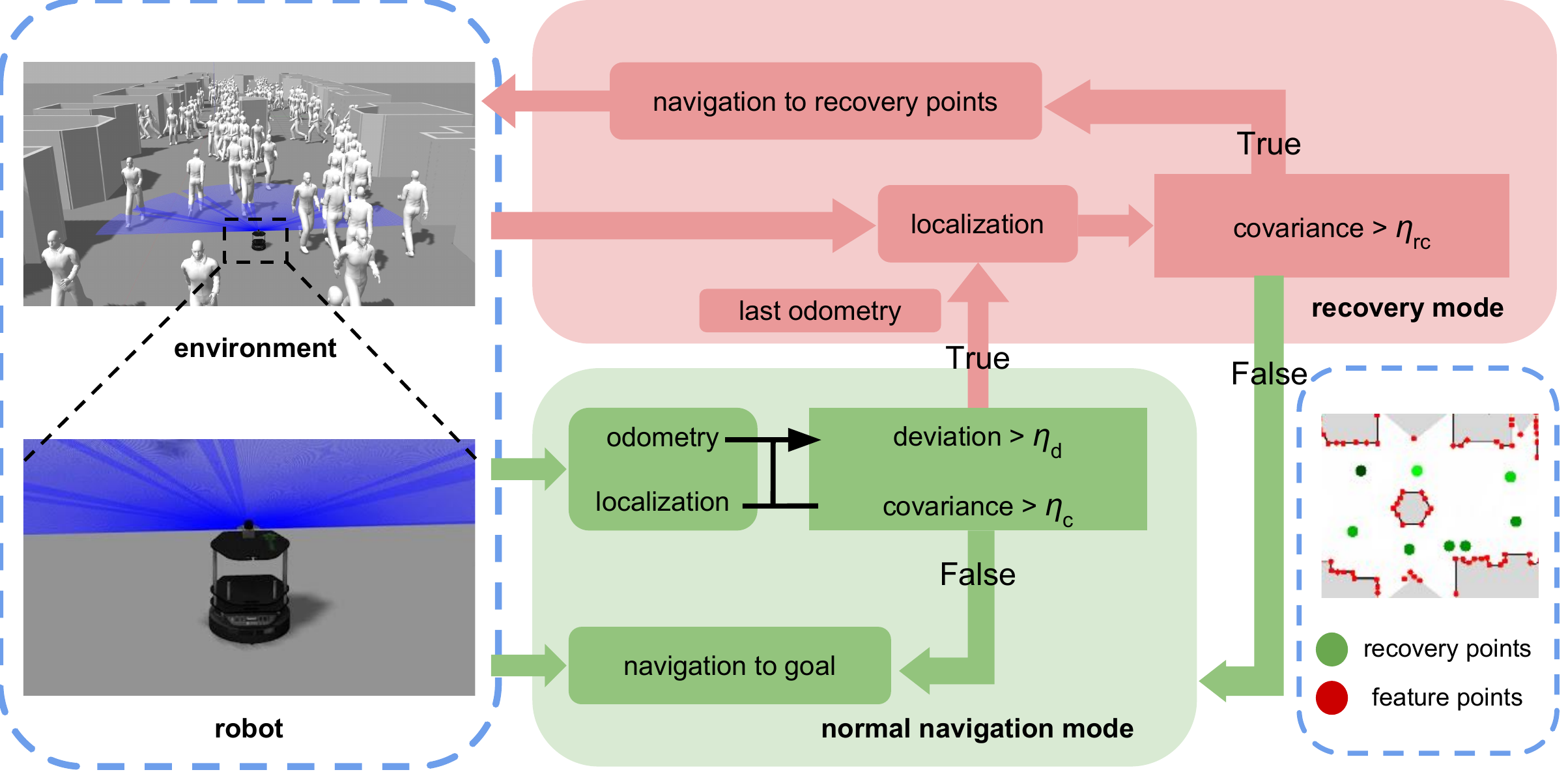}
\caption{The architecture of our navigation system for getting a robot unfrozen and unlost in a complex scenario with both static obstacles and moving pedestrians. The navigation policy is mainly composed of two parts: a normal navigation policy that can effectively avoid obstacles and a localization recovery policy that can help the robot to accurately localize itself in the global map.}
\label{fig:overview}
\end{figure*}
In addition, the \textit{robot freezing} problem and the \textit{navigation lost} problem are tightly coupled in a dense crowd, which usually is not considered in prior works. On the one hand, given a plan about ``where to move'' and ``where to look'' for resolving the \textit{navigation lost} problem, the robot needs to accomplish the actual movement in the physical world. To make such movement safe and collision-free, the robot must have high mobility in the dense crowds, which requires solving the \textit{robot freezing} problem as a prerequisite. On the other hand, before executing the actions toward the goal for ``moving'' or ``looking'', the navigation algorithm needs to understand the routes between the robot's current position and the target location, which requires an accurate localization of the robot in the map, i.e., we need to solve the \textit{navigation lost} problem first.

In this paper, we solve the challenging problem of robotic global navigation in crowd scenarios. In particular, given a map and a goal, the robot needs to navigate through the dense crowds and complex static obstacles and eventually reach the goal safely, accurately, and efficiently without getting frozen or lost. This task can be applied in many real-world scenarios. We present a novel framework to handle the \textit{robot freezing} and the \textit{navigation lost} problems simultaneously. As illustrated in \prettyref{fig:overview}, our framework consists of two modes: the \textit{normal mode} and the \textit{recovery mode}. 
In the normal mode, our robot is driven by a LiDAR-based localization algorithm and a reinforcement learning based local planner, which work together to endow the robot with the goal-approach ability as well as excellent collision-avoidance mobility. 
However, the normal navigation policy can hardly tackle the navigation lost problem that is ubiquitous in situations with extremely high-density. More specifically, as the robot navigates with the dense crowds, the LiDAR-based localization will easily fail and then the robot will lose the knowledge about where it is.  In this situation, the robot will switch into the localization recovery mode. In particular, we sample a set of locations with rich landmark features, called recovery points, from a given 2D map. The robot will dynamically select one of these discriminative points and then approach the selected point to re-localize itself in the map. The optimal recovery point is determined using a reinforcement learning-based optimization that maximizes the accessibility of the recovery point, in order to optimally balance the benefit and overhead of the recovery action. During the navigation, the robot will actively switch between the normal mode and the recovery mode, according to the instantaneous situation in the scenario. 
To assess the performance of our proposed algorithm in such a challenging navigation task, we set up a set of complex simulated environments as the testing benchmark, which will be released upon the publication of this paper.
We also design a set of metrics for performance evaluation. The experimental results verify the excellent mobility and reliability of our method in challenging navigation scenarios.

Our contributions can be summarized as follows:
\begin{itemize}
\item We address for the first time the challenging problem of the robot navigation in complex environments through dense crowds without suffering from getting frozen or getting lost. 
\item we formulate a novel framework to handle the \textit{navigation lost} and the \textit{robot freezing} problem simultaneously, by switching between the normal mode and the recovery mode during navigation.
\item We propose a reinforcement learning based recovery algorithm that enables the robot to regain the localization by approaching recovery points adaptively.
\item We provide a benchmark including simulated and real-world scenarios to evaluate navigation algorithms and to demonstrate our proposed method's superior performance. 
\end{itemize}

\section{Related Work}
\label{sec:related}

The \emph{robot freezing} problem has been widely studied for mobile robot navigation. 
One ad-hoc solution to resolve the suboptimal frozen state is to follow an essentially arbitrary path through the crowd, but such highly evasive paths often are dangerous and thus are not desirable for service robot applications. One culprit behind the freezing robot problem is the uncertainty explosion, i.e., the union of the overly conservative predictions about the trajectories of nearby moving obstacles blocks all the possible movements of the robot and makes the robot fail to find a clear path passing through the dense crowd. Some previous research thus focused on controlling the predictive covariance, for instance, by repetitive re-planning~\cite{Otte:2016:RRT,Berg:2006:ICRA}, belief space feedback planning~\cite{VanDenBerg:2012:MPU}, or developing high-fidelity independent human movement models~\cite{Alahi:2016:CVPR,Yi:2016:ECCV,Gupta:2018:CVPR,Kim:2015:BRVO}. However, as argued in~\cite{Trautman:2010:IROS,Trautman:2013:ICRA,trautman2015robot}, even perfect individual prediction (i.e., when the robot is aware of all other agents’ accurate trajectories) cannot get rid of the freezing robot problem when the navigation algorithm is lack of the mathematical models of cooperation between the robot and humans. This is because when the robot is not anticipating cooperation, it will still choose a highly evasive maneuver rather than adapting its trajectory to the humans to make room for navigation. As a result, it is concluded that a model for joint collision avoidance among nearby agents is a prerequisite for effective navigation in the dense crowd, which unfortunately is not available in most of the previous navigation algorithms.

The \emph{navigation lost} problem arises when the robot's localization uncertainty accumulated during the navigation becomes so large that the robot cannot accurately locate itself in a given map~\cite{Fox:1999:MCL}. The robot may also get lost due to the localization error, e.g., in a dynamic environment, the robot may mix up known static obstacles with unknown dynamic obstacles~\cite{Fox:1999:MLM}. Most previous solutions to the navigation lost problem are passive methods. They assume that the robot motion and the pointing direction of the sensors cannot be controlled, and focus on selectively utilizing the sensor stream to minimize the localization uncertainty or error, e.g., by using different filters~\cite{Fox:1999:MLM} or more sophisticated modeling about the dynamic scenes~\cite{Tipaldi:2013:LLC,Sun:2016:IROS}. However, for highly dynamic scenarios with dense human crowds, the pedestrians may block all the landmark features necessary for localization, and thus the robot must choose sophisticated policies for determining the robot's motion direction and the camera's pointing direction. Some methods recover the localization by asking the robot to look at places with special properties, e.g., with high saliency~\cite{Siagian:2009:BIM}, road tracks~\cite{Chang:2013:Beobot}. Some other approaches try find optimal actions that can minimize metrics with respect to the localization quality, including the entropy~\cite{Burgard:1997:AMR} or number of hypotheses about the robot's current location~\cite{Li:2016:ALD}. The localization recovery problem can also be formulated and solved under the more general POMDP (Partial Observable Markov Decision Making) framework, which can be solved in a tractable manner by using Gaussian belief space approximations~\cite{VanDenBerg:2012:MPU,Platt:RSS:BSP}.


\section{Proposed Framework}
To tackle the robot freezing problem and the navigation lost problem simultaneously, as illustrated in \prettyref{fig:overview}, we present a novel navigation framework that controls the robot's behavior in two modes: the \emph{normal navigation mode} and the \emph{localization recovery mode}. During navigation in challenging scenarios, the robot will actively switch between these two modes in an online manner to accomplish the navigation task. 

\subsection{Normal navigation policy}
\label{sec:normal}
In the normal mode, the robot utilizes the SLAM algorithm to accomplish the normal navigation mission, e.g., approaching its goal in (un)structured static environments. In particular, we use the state-of-the-art LiDAR SLAM algorithm, Cartographer~\cite{Cartographer}, as our basic localization module. In general, Cartographer can handle the low-density dynamic obstacles based on the map information updated in real-time. 
However, as the density of dynamic obstacles (such as human crowds) increases, its performance can be degraded significantly. Therefore we first incorporate a reinforcement learning based collision avoidance method, which can significantly improve the robot's mobility to approach the goals compared with the traditional local planner when avoiding collisions in the dense crowds. In particular, as in our previous work~\cite{long2017towards}, we used an Actor-Critic based PPO algorithm~\cite{ppo} to train a local planner for crowd avoidance. The Actor-Critic framework ~\cite{konda2000actor} has been widely used in the reinforcement learning scheme. Commonly, the Actor module serves as the controller, while the Critic module is used to guide the gradient update of the Actor module by minimizing the surrogate loss
\begin{equation}
L(\theta) = \hat{\mathbb{E}}_t [\text{min} (r_t(\theta)\hat{A}_t, \text{clip}(r_t(\theta), 1 - \epsilon, 1+\epsilon)\hat{A}_t)],
\label{eqn:policy_gradient}
\end{equation}
where $\hat{A}_t$ is an estimation to the advantage function, $r_t$ is the reward at timestep $t$, $r_t(\theta)$ denotes the probability ratio $r_t(\theta) = \frac{\pi_{\theta}(a_t \mid s_t)}{\pi\textsubscript{old}(a_t \mid s_t)} \hat{A}_t$, and $\epsilon$ is a hyperparameter. $\hat{A}_t$ is computed as \begin{equation}
\hat{A}_t = {\delta}_t + ({\lambda}{\gamma}){\delta}_{t+1} + \dotsb + ({\lambda}{\gamma})^{T-t+1}{\delta}_{T-1},
\label{eqn:advantage_fun}
\end{equation}
where ${\delta}_t = r_t + {\gamma}V_{\pi}(s_{t+1}) - V_{\pi}(s_t)$, with ${\gamma}^t$ as the discounting factor ${\gamma} \in [0, 1]$ and $V_{\pi}(s_t)$ as the state value function for the state $s_t$, and 
\begin{equation}
V_{\pi}\left ( s_{t} \right ) = \mathbb{E}\left [ R \right ] =\mathbb{E}\left [ \sum_{t=0}^{\infty }\gamma^{t}r_{t}\right ].
\label{eqn:value_fun}
\end{equation}

In this paper, we train the Actor and Critic in a way similar to our previous paper~\cite{long2017towards}. The Actor and Critic's inputs are the robot's laser scan $\mathbf{s}\textsubscript{scan}$, the velocity of robot itself $\mathbf{s}\textsubscript{vel}$, and goal information $\mathbf{s}\textsubscript{goal}$, while the output is a collision-free velocity command fed to the robot. It is worth noting that here the RL-based collision avoidance policy plays a different but more important role than the one as a local motion planner in our previous work~\cite{long2017towards}. Inspired by the work~\cite{faust2017prm}, here the RL-based collision avoidance is used to accomplish a global planner by combining with traditional grid-based global planners. In particular, the goal information $\mathbf{s}\textsubscript{goal}$ now is the sub-goal assigned by the grid-based global planner rather than the final goal of the agent. 





\begin{figure*} [t]
\centering
\includegraphics[width=1\linewidth]{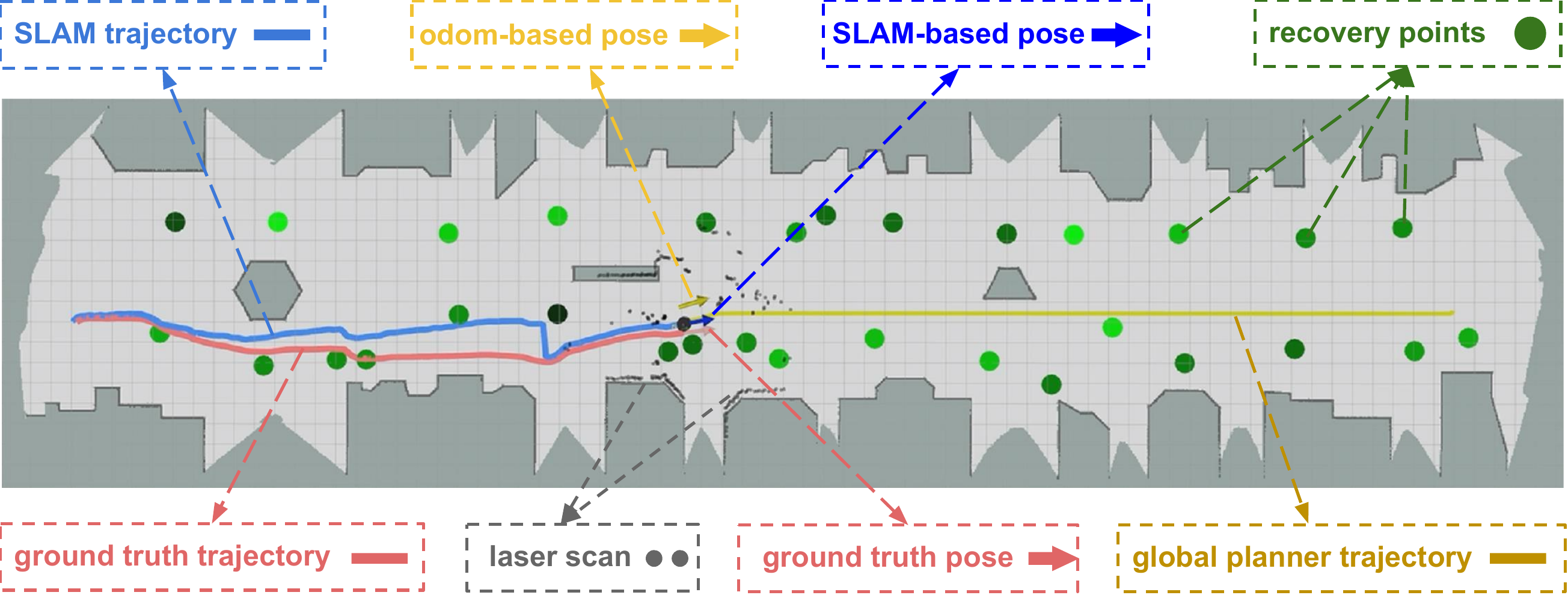}
\caption{System inputs and outputs of our navigation system, visualized by Rviz \protect  \footnotemark. }
\label{fig:fig_detail}
\end{figure*}

\subsection{Localization recovery policy}
The localization could get lost during navigation in a set of challenging situations, such as in featureless places, when sensor views being blocked, or when the robot is getting stuck in the middle of a dynamic crowd. 
To deal with these cases, we present a recovery mechanism so that the robot can adaptively switch to the recovery mode and regain the localization certainty. 
We first precompute a set of candidate recovery points in the global map, which can facilitate the localization recovery. Then, we use a reinforcement learning based recovery policy to determine to approach which recovery point for localization recovery. Note that this policy will continuously select a suitable recovery point during the navigation and it may switch to another recovery point before reaching the recovery point chosen last time, to optimally balance navigation efficiency and localization uncertainty.

\subsubsection{Candidate recovery points}
Given the 2D grid map obtained offline from Cartographer~\cite{Cartographer}, we need to recognize locations that are more helpful for re-localization. 
One straightforward solution is to step back to the starting point and then find another way to achieve the destination. This policy, even though sometimes adopted by humans, however, is not efficient and will suffer from the freezing problem when the scenario is crowded. 
Here, similar to the visual tracking and visual SLAM, we choose the recovery points from regions with sharp corners or fine structures, whose spatial invariance and stability of matching are beneficial for re-localizing the robot in the global map. In particular, we use the Harris corner detector~\cite{harris1988combined} to extract corners from the 2D map. Next, we use $K$-Means to cluster the extracted corners. To refine the clustering, if any corner is found far from its cluster centroid, a new cluster will be created, which naturally prevents corners far away from being clustered together. 
Since the cluster centroids may not locate in the passage-ways of the map (e.g., it may be close to the boundaries of the environment), we offset the centroids towards the map center into the passage-ways and the corresponding final positions are treated as candidate recovery points, as shown in the bottom-right in \prettyref{fig:overview}. To facilitate the recovery point selection, we assign each of them a preference weight $v_{cp}^{i}$:
\begin{equation}
v_{cp}^{i} = \frac{N_{cp}^{i}}{\sum_{k=1}^{n}N_{cp}^{k}},
\label{eqn:map_values}
\end{equation}
where $N_{cp}^{i}$ is the number of corners in the $i$-th cluster. Thus $v_{cp}^{i}$ is proportional to the number of detected corners belonging to the same cluster. In this way, the recover cluster with more corner points will be preferred. This is desirable for the robustness of the recovery. In particular, the robot's localization uncertainty will continue increasing when approaching the recovery region and thus it cannot reach the recovery point exactly. If the recovery region has only a few corners, the robot may miss all of them due to the localization error and the localization recovery will fail. If the region has many corners, the robot's recovery task has a higher probability to succeed.




\footnotetext{{\url{http://wiki.ros.org/rviz}}}

\subsubsection{Actor-Critic based recovery}
After we compute the candidate recovery points, we hope that the recovery algorithm can automatically choose a near-optimal recovery point by combining the knowledge about the number of features near the recovery points, the flow of surrounding dynamic obstacles (e.g., pedestrians), and the distance between the recovery points and the eventual destination.

\begin{figure}[!htb]
\centering
\includegraphics[width=1.0\linewidth]{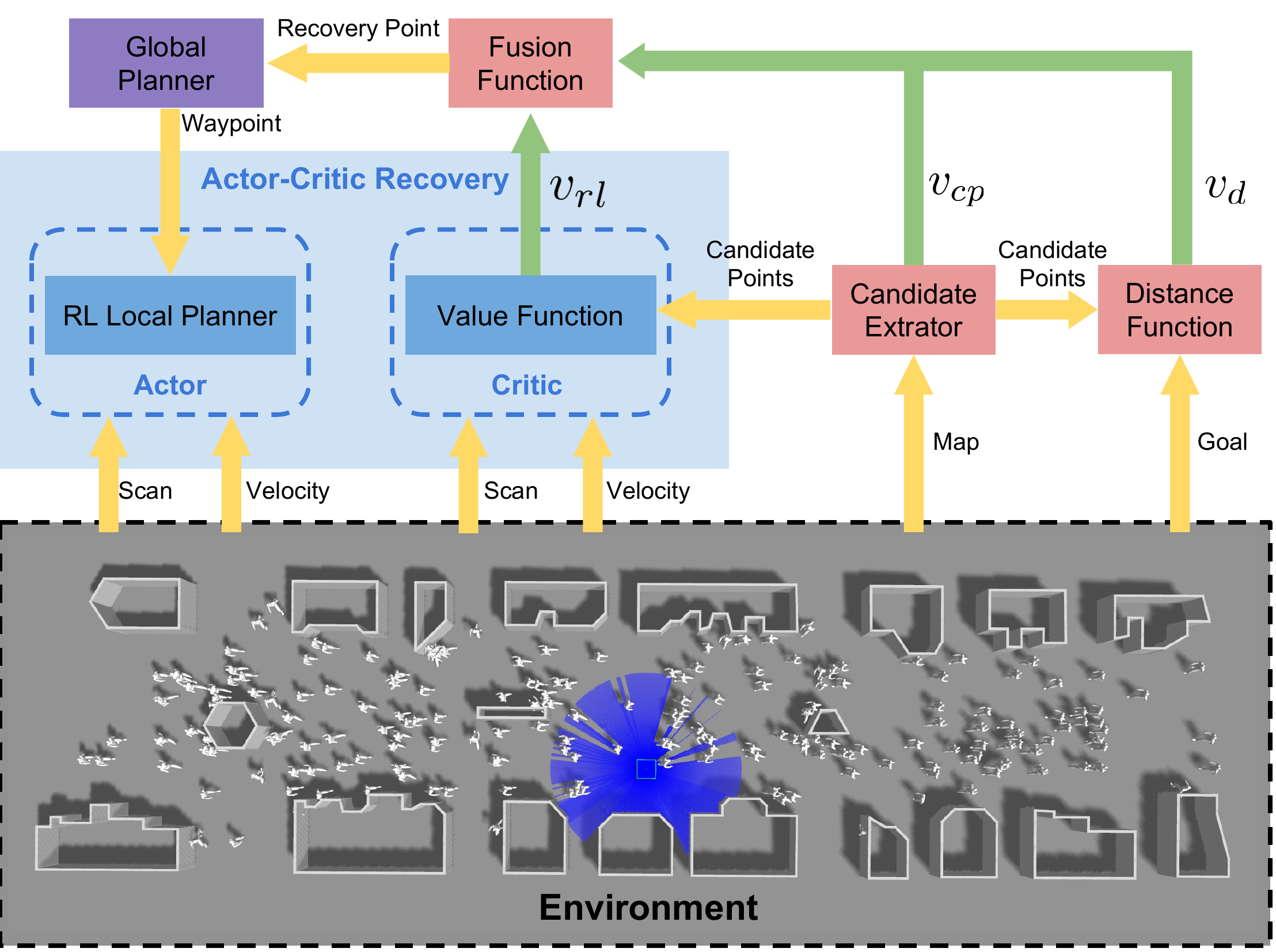}
\caption{Actor-Critic based localization recovery policy.}
\label{fig:rl_recovery_behavior}
\end{figure}

In order to estimate the chance of the robot reaching one candidate recovery point under the surrounding dynamic environment, we adopt the Actor-Critic framework. Different with the Actor-Critic for normal navigation in \prettyref{sec:normal}, here the Critic no longer guides the gradient update of the Actor, but guides the Actor to the most accessible recovery point under current dynamic situation. In other words, we utilize the Critic module as a high-level guidance for recovery point selection, while the Actor module is implemented using our collision avoidance method for the low-level planning policy. Although the Critic is used for different purposes compared to that in \prettyref{sec:normal}, we do not retrain a Critic for the recovery policy, but instead using the value function in \prettyref{eqn:value_fun} as a Critic.
This is a reasonable solution because when we used reinforcement learning algorithm to train Actor and Critic, the reward function is designed in such a way that the robot will obtain a high reward when it reaches the goal assigned by the user. Thus, according to ~\prettyref{eqn:value_fun}, Critic will learn to give higher scores to those world states in which the robot is more likely to reach a given target. In this way, the Critic can use the laser scan data to evaluate the chance of passing through the surrounding dynamic obstacles to reach each destination. Similarly, in the recovery policy, we can use Critic to evaluate the accessibility of a recovery point according to the current observation of the pedestrian flow, formally implemented as
\begin{equation}
v^i_{rl}(\mathbf{p}^i_{\text{recovery}}) = V_{\pi}(\mathbf{s}\textsubscript{scan}, \mathbf{s}\textsubscript{vel}, \mathbf{p}^i_{\text{recovery}}),
\label{eqn:values_rl}
\end{equation}
where $V$ is the value function of the Critic, and $\mathbf{p}_{\text{goal}}^{i}$ is the position of the $i$-th recovery point. In other words, the higher the value the Critic given, the easier for the robot to reach the recovery point, and vice versa. 

In addition, we also prefer a recovery point that is on the robot's way toward the final destination, because the recovery points that are off the way would suffer the navigation efficiency. Thus, we define another evaluation function $v^i_{d}(\cdot)$  for the recovery points as:
\begin{equation}
v^i_{d}(\mathbf{p}^i_{\text{recovery}}) = \frac{-\| \mathbf{p}\textsubscript{goal} - \mathbf{p}^i_{\text{recovery}}\|_2}{\sum_{i=1}^n \| \mathbf{p}\textsubscript{goal} - \mathbf{p}^i_{\text{recovery}} \|_2},
\label{eqn:values_goal}
\end{equation}
where $\mathbf{p}\textsubscript{goal}$ is the position of the final goal, and $\mathbf{p}^i_{\text{recovery}}$ again is the position of the $i$-th recovery point.

Finally, to select the most suitable point for recovery, we consider a set of different factors, including the preference weight of a candidate recovery points, the distance from each candidate to the goal, and the information about the dynamic environment, i.e., the pedestrian flow in the scenario. Formally, the optimal $k^*$-th recovery point is determined as
\begin{equation}
k^* = \arg\max_{k} \Big( \omega _{rl}\cdot v^k_{rl} + \omega _{cp}\cdot v^k_{cp} + \omega _{d}\cdot v^k_{d} \Big),
\label{eqn:fusion_fun}
\end{equation}
where $\omega_{rl}$, $\omega_{cp}$, and $\omega_{d}$ are hyperparameters. 

After deciding which recovery point to approach, the chosen recover point is treated as an intermediate goal and is fed to the global planner as a waypoint. Then the global planners passes the generated sub-goal to the reinforcement learning based local planner that we developed in~\cite{long2017towards} for local navigation. The entire localization recovery process is shown in \prettyref{fig:rl_recovery_behavior}.

\subsection{Switch strategy}
\label{sec:switch}

In previous sections, we have described two modes of our navigation policy in detail.
Here we will explain how to switch between these two modes automatically. We propose two simple yet efficient trigger conditions for switching, specifically, the deviation between odometry and SLAM localization and the covariance for SLAM localization. We believe that the odometry position system may be not accurate enough due to the slipping and cumulative error, but it will not drift a large distance in an instant. To the contrary, the SLAM positioning system is generally accurate  but may have a significant drift if the mistake happens during the feature matching between the laser data and the global map. Thus the difference between the outputs of these two localization systems can be used as the signal for the mode switch. We also the covariance for SLAM localization output to track the uncertainty of SLAM system. If any of these two values is larger than a given threshold, the system will step into the \emph{recovery mode} automatically. 
As shown in \prettyref{fig:overview}, while entering the recovery mode, the robot will first start recovering from the last odometry output. Next, it will move to the selected recovery point in an adaptive manner. The robot will get back to the \emph{normal mode} when the recovery procedure succeeds, i.e., when the covariance of the robotic localization is smaller than a given threshold.

\section{Experiments and Results}
\label{sec:exp}

\begin{figure*}[h]
\centering
\includegraphics[width=1\linewidth]{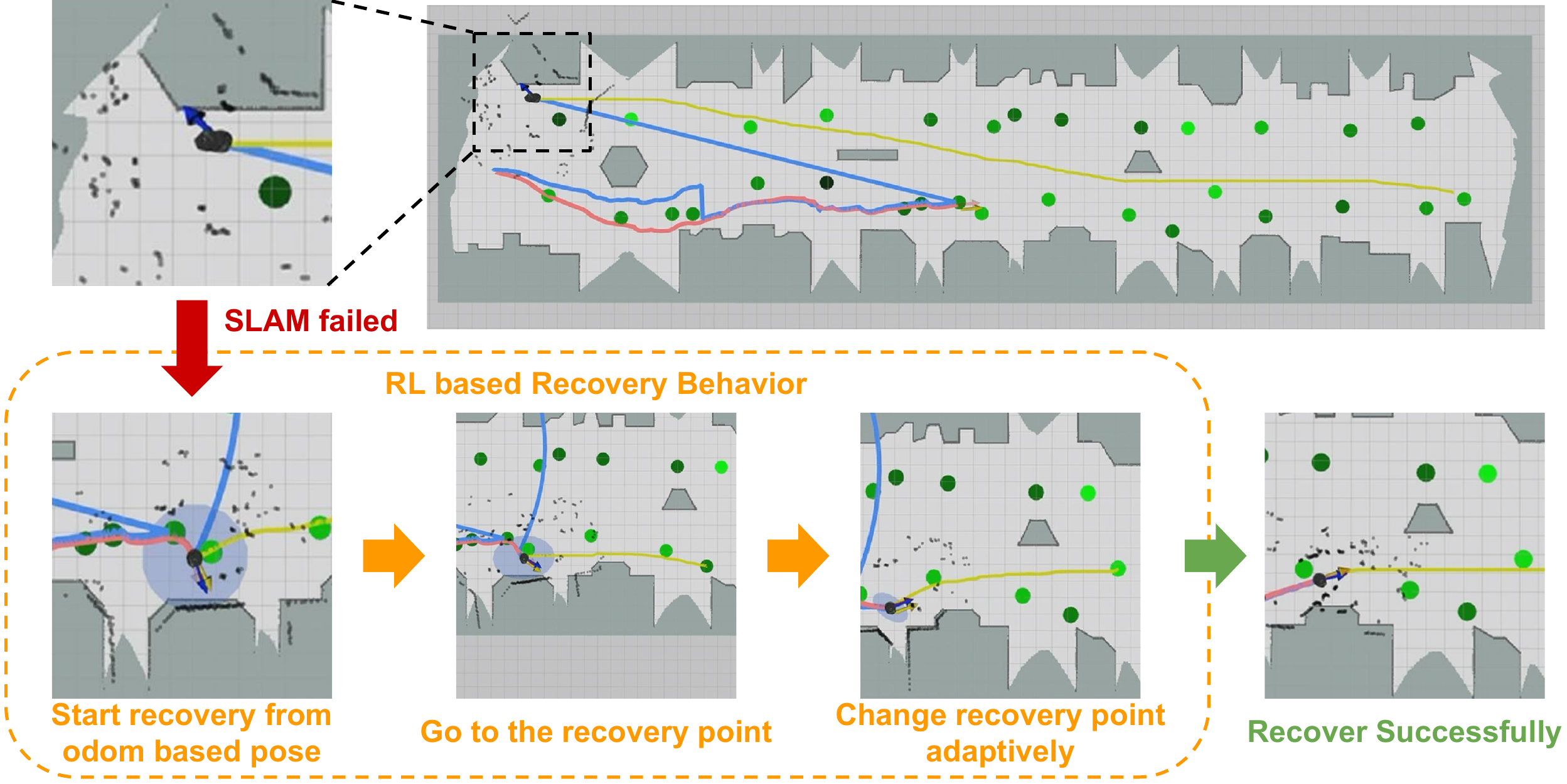}
\caption{The demonstration about how the reinforcement learning based localization recovery policy helps the robot  to re-localize itself from a SLAM failure in a step-by-step manner.}
\label{fig:recovery}
\end{figure*}

In this section, we first briefly introduce our simulation platform and define the evaluation metrics.
Then we introduce the details of the entire simulation experiment. Finally, we analyze our experimental results and run the algorithm successfully in a real robot.

\subsection{Experiment setup and metrics}
\begin{figure*}[htb]
\centering
\begin{subfigure}{0.32\textwidth}
\centering
\includegraphics[width=1\linewidth]{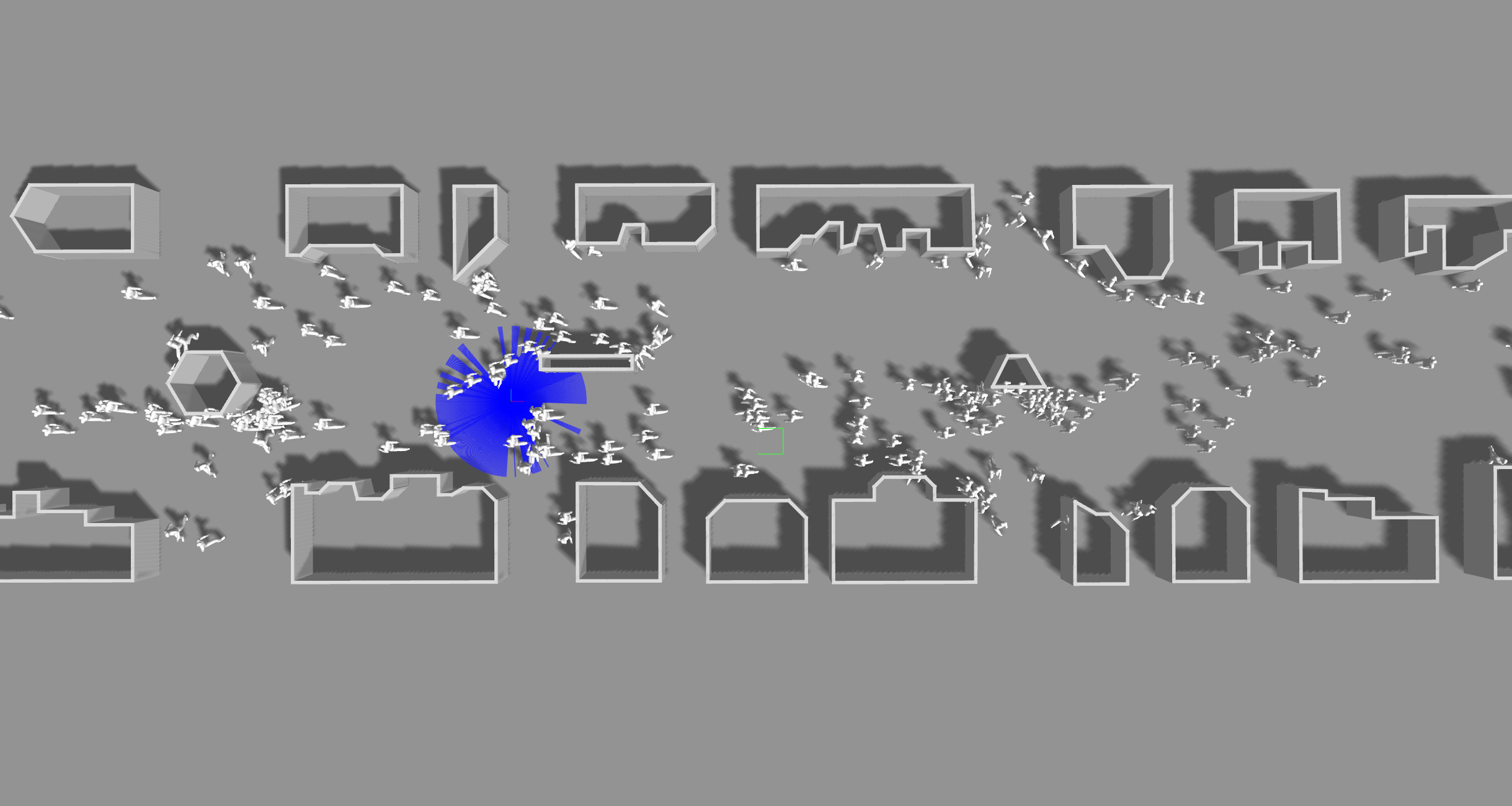}
\caption{corridor environment}
\label{fig:corridor_env}
\end{subfigure}
\begin{subfigure}{0.32\textwidth}
\centering
\includegraphics[width=1\linewidth]{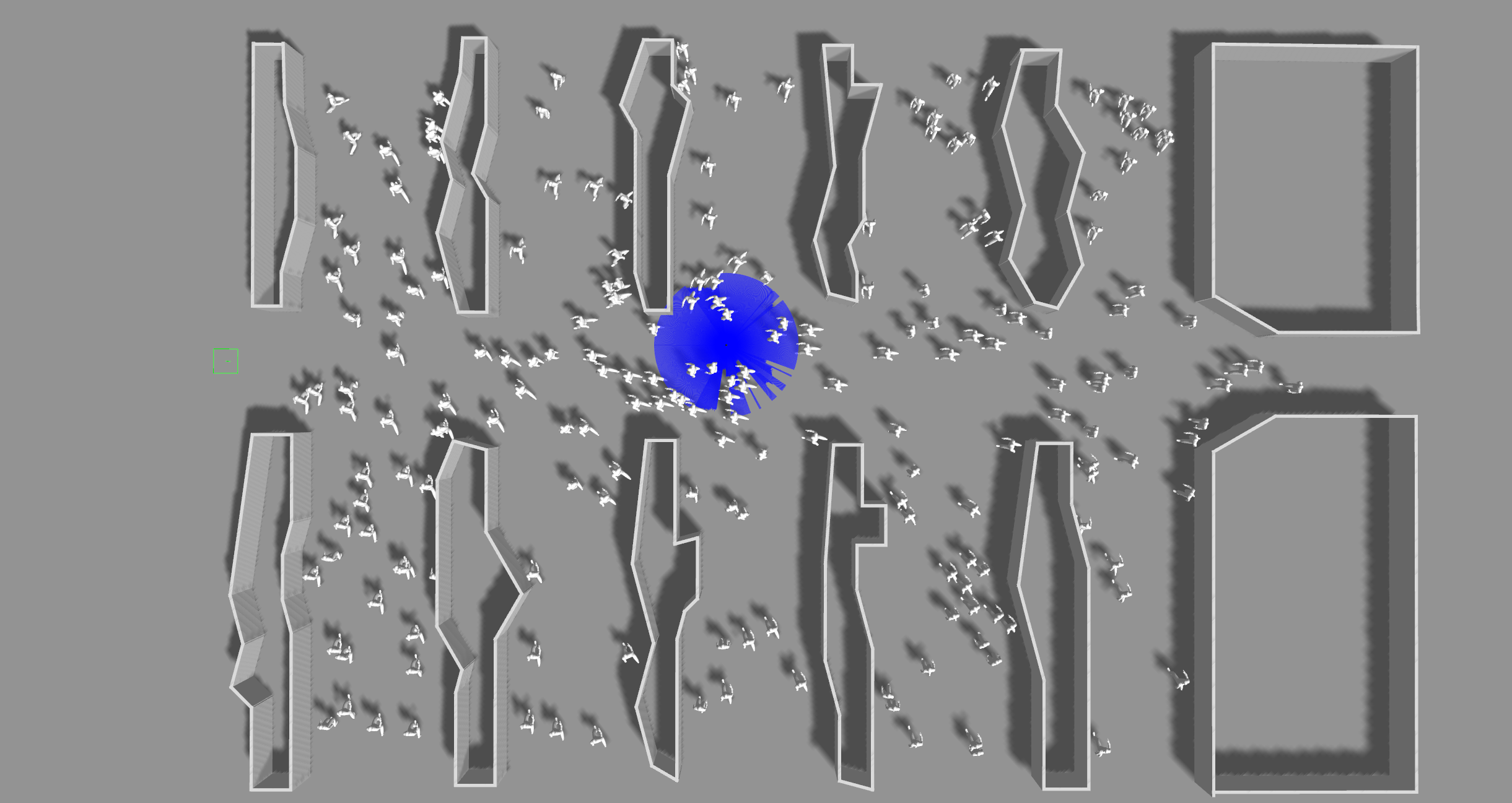}
\caption{supermarket environment}
\label{fig:supermarket_env}
\end{subfigure}
\begin{subfigure}{0.32\textwidth}
\centering
\includegraphics[width=1\linewidth]{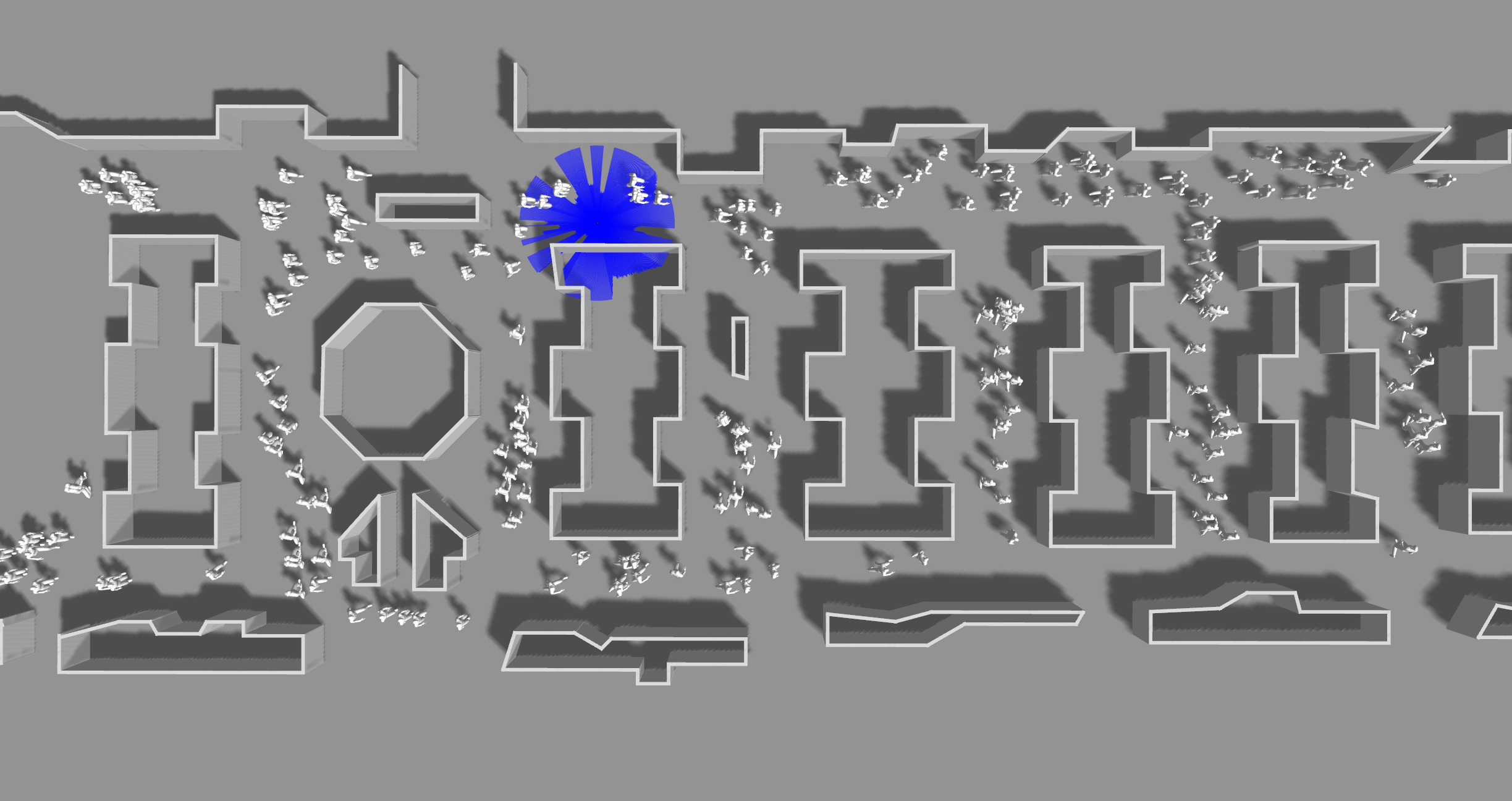}
\caption{airport environment}
\label{fig:airport_env}
\end{subfigure}
\caption{The three simulation scenarios used as the benchmark for testing navigation performance.}
\label{fig:sim_env}
\end{figure*}

To build a simulation environment for quantitatively measuring the navigation effectiveness in a complex dynamic environment, we combine two simulators, Gazebo\footnote{\url{http://gazebosim.org/}} and Menge~\cite{menge}.
The Gazebo simulator is responsible for the simulation of the robot system, while Menge serves as the driver for simulating the crowd movement within the Gazebo environment. 
Then, we build three typical scenarios including the corridor, the supermarket, and the airport, all with crowds as shown in \prettyref{fig:sim_env}. We use the Turtlebot2 as our mobile robot platform. According to the actual hardware parameters of the Rplidar A2 sensor mounted on the real robot, for the 2D LiDAR sensor in the simulation we set its sweep range to \ang{360}, the maximum sweep radius to \SI{6}{m}, and the angular resolution to \ang{1} per range, which is further added up with the Gaussian white noise with 0.05 standard deviation.

As described in \prettyref{fig:fig_detail}, we can estimate the position of the robot in two ways, either from odometer or from the Cartographer localization algorithm. And from the simulator, we have access to the ground truth of the robot position, also shown in \prettyref{fig:fig_detail}. 
Given the mesh model of a simulation scenario, we can also compute the set of candidate recovery points, which are colorized according to the preference weights calculated by  \prettyref{eqn:map_values}, where the lighter the color, the greater the weight.

To verify that our algorithm can effectively reduce the probability of robots getting lost or frozen in the crowd, we propose three metrics to evaluate the navigation performance, i.e., the \textit{lost rate}, the \textit{frozen rate}, and the \textit{success rate}. We define a robot to be lost if the distance between the SLAM estimation of the robot position $\mathbf{p}\textsubscript{estimate}$ and the ground truth position of the robot $\mathbf{p}\textsubscript{robot}$ from the simulator is always greater than a given lost threshold $\mathbf{d}\textsubscript{lost}$ in a period of $\Delta t\textsubscript{lost}$. More formally, the robot is
\begin{equation}
\begin{cases}
\text{lost} & \quad \text{if } \left \| \mathbf{p}\textsubscript{robot} - \mathbf{p}\textsubscript{estimate} \right \|_{t} > \mathbf{d}\textsubscript{lost}, \forall t\in [ 0,  \ \Delta t\textsubscript{lost}]   \\ 
\text{unlost} & \quad \text{otherwise}.  \\
\end{cases}
\label{eqn:lost_rate}
\end{equation}

Because a robot with the frozen issue will stop moving forward and just turn around in place, we determine whether a robot is frozen or not based on whether its linear speed along the robot's forward direction $v^\perp_{\text{robot}}$ is always no less than a given threshold speed $v\textsubscript{frozen}$ during the time period $\Delta t\textsubscript{frozen}$. More formally, the robot is
\begin{equation}
\begin{cases}
\text{frozen} & \quad \text{if } v^\perp_{\text{robot}} < v\textsubscript{frozen}, \forall t\in [ 0,  \ \Delta t\textsubscript{frozen}]\\ 
\text{unfrozen} & \quad \text{otherwise}.  \\
\end{cases}
\label{eqn:freezing_rate}
\end{equation}

We determine whether a robot succeeds in the navigation task by checking the robot's arrival at the target position, i.e., whether the robot has 
\begin{equation}
\begin{cases}
\text{arrived} & \quad \text{if } \| \mathbf{p}\textsubscript{robot} - \mathbf{p}\textsubscript{goal} \| < r\textsubscript{arrive} \\
\text{not arrived} & \quad \text{otherwise},  \\
\end{cases}
\label{eqn:success_rate}
\end{equation}
where $\mathbf{p}\textsubscript{goal}$ is the goal position and $r\textsubscript{arrive}$ is the threshold for arriving.

Based on \prettyref{eqn:lost_rate}, \prettyref{eqn:freezing_rate}, and \prettyref{eqn:success_rate}, we can compute the lost rate, the frozen rate, and the success rate of the navigation algorithm. 

In addition, we also want to investigate whether our algorithm will have an impact on the navigation efficiency of the robot in terms of the time cost reaching the target point. Thus, we further evaluate the mean robot speed for the navigation trials in which the robot successfully reaches the destination.

\subsection{Implementation details}
In this part, we first summarize in \prettyref{tab:hyperparameters} the hyperparameters used in our algorithm. 


\begin{table}[h]
\centering
	\caption{Hyperparameters used in our system}
	\label{tab:benchmark}
	\fontsize{9.5}{9.5}\selectfont
	\bgroup
	\def\arraystretch{1.3}
 \begin{tabular}{lllll} 
\hline
Parameter & Value &  & Parameter & Value  \\ 
\hline
\text{$\omega_{rl}$ in \prettyref{eqn:fusion_fun}}        	& 0.5    &  & \text{$\mathbf{d}\textsubscript{lost}$ in \prettyref{eqn:lost_rate}}        & 3.0     \\
\text{$\omega_{cp}$ in \prettyref{eqn:fusion_fun}}        	& 0.2    &  & \text{$\Delta t\textsubscript{lost}$ in \prettyref{eqn:lost_rate}}        & 10    \\
\text{$\omega_{cp}$ in \prettyref{eqn:fusion_fun}}        	& 1.0    &  & \text{$v\textsubscript{frozen}$ in \prettyref{eqn:freezing_rate}}        & 0.2     \\
\text{$\eta_{}$ in \prettyref{fig:overview}}	          	& 3.0    &  &  \text{$\Delta t\textsubscript{frozen}$ in \prettyref{eqn:freezing_rate}}     & 10     \\
\text{$\eta_{rc}$ in \prettyref{fig:overview}} 		        & 0.08    &  & \text{$r\textsubscript{arrive}$ in \prettyref{eqn:success_rate}}        & 0.5   \\
\text{$\eta_{c}$ in \prettyref{fig:overview}}		       & 0.2   &  &           &        \\
\hline
\end{tabular}
\egroup
\label{tab:hyperparameters}
\end{table}

Then we make a comprehensive comparison among three different approaches (baseline, RL, and (RL)$^2$) in all the three testing scenarios as shown in \prettyref{fig:sim_env}. The baseline method 
combines the ROS movebase navigator with the Cartographer localization~\cite{Cartographer}, where the movebase navigator uses the 
dynamic window approach~\cite{fox1997dynamic} for local planning and Dijkstra algorithm for global planning. The RL method replaces the local planner in the baseline method with the deep reinforcement learning based local planner~\cite{long2017towards}. The (RL)$^2$ method is the approach that we proposed in this work, which uses the reinforcement learning for both the local collision avoidance and localization recovery.


\subsection{Result analysis}
\prettyref{tab:benchmark} shows the comparison results on three different scenarios in our benchmark. When using the baseline navigation policy, the robot is almost impossible to reach the goal due to the high density and few features in the dense crowd scenario. When using the RL navigation policy, the frozen rate declines significantly and the robot gets some chance to reach the goal, thanks to the high mobility of the RL-based local planner. However, this method does not deal with the lost problem and thus the robot can get lost in the scenario. The (RL)$^2$ policy can significantly increase the success rate of the navigation task because it considers the lost and frozen issues simultaneously. One interesting phenomenon is that the frozen rate of (RL)$^2$ is higher than that of the RL in the supermarket and the airport scenarios. This is because the RL policy focuses on the local collision avoidance and does not pay attention to the re-localization. Thus its risk of getting stuck is lower. However, it has a much higher risk in getting lost in the large scenarios like the supermarket and the airport, which is supported by RL's much larger lost rate than that of (RL)$^2$.

\begin{table*}[htb]
	\centering
	\caption{Comparison of the navigation performance of different navigation policies in three scenarios.}
	\label{tbl:benchmark}
	\fontsize{8.5}{8.5}\selectfont
	\bgroup
	\def\arraystretch{1.3}
	\setlength{\tabcolsep}{3.5pt} 
    \begin{tabular}{llcccclcccclcccc} 
    \toprule
    \multicolumn{1}{c}{\multirow{2}{*}{Methods}} &  & \multicolumn{4}{c}{Corridor Scenario}           &  & \multicolumn{4}{c}{SuperMarket Scenario}        &  & \multicolumn{4}{c}{Airport Scenario}             \\ 
    \cline{3-6}\cline{8-11}\cline{13-16}
    \multicolumn{1}{c}{}                         &  & Lost & Frozen & Success       & Velocity      &  & Lost & Frozen & Success       & Velocity      &  & Lost & Frozen & Success       & Velocity       \\
    baseline                      &  & 32\% & 65\%     & 3\%           & 0.52          &  & 18\% & 82\%     & 0\%           & 0             &  & 16\% & 84\%     & 0\%           & 0              \\
    RL                     &  & 63\% & 17\%     & 20\%          & \textbf{0.76} &  & 86\% & 10\%     & 4\%           & \textbf{0.81} &  & 71\% & 5\%      & 24\%          & \textbf{0.68}  \\
    (RL)$^2$                                         &  & 36\% & 3\%      & \textbf{61\%} & 0.73          &  & 20\% & 12\%     & \textbf{68\%} & 0.78          &  & 10\% & 28\%     & \textbf{62\%} & 0.60           \\
    \bottomrule
    \end{tabular}
	\egroup
	\label{tab:benchmark}
\end{table*}

The superiority of (RL)$^2$ in performance is due to our novel recovery policy. In particular, when the robot realizes that it may get lost, the response of the baseline policy is to turn around in place for safety and look for nearby features to rescue the robot from the lost, which can be an efficient solution in a static or moderately dynamic scene. However, in a highly dynamic scenario, even though its lost level will not get worse, the robot will get frozen and thus still cannot reach the goal successfully. While using our recovery policy, rather than simply turning around in place, the robot will first find a path to break free of the crowded neighborhood and then move toward an accessible recovery point to re-localize itself. Once the robot re-localizes successfully, it will switch back to the normal navigation mode. \prettyref{fig:recovery} provides a step-by-step illustration of this procedure.

\begin{figure}[h] 
\centering
\begin{subfigure}{0.23\textwidth}
\includegraphics[trim=10 0 10 0, clip, width=1.0\linewidth]{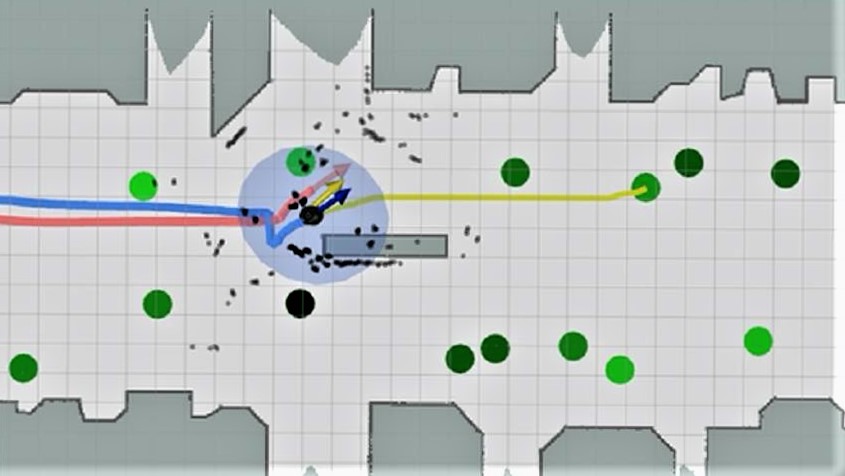}
\caption{ }
\label{fig:adaptive_a}
\end{subfigure}
\begin{subfigure}{0.23\textwidth}
\includegraphics[trim=10 0.5 10 0.5, clip, width=1.0\linewidth]{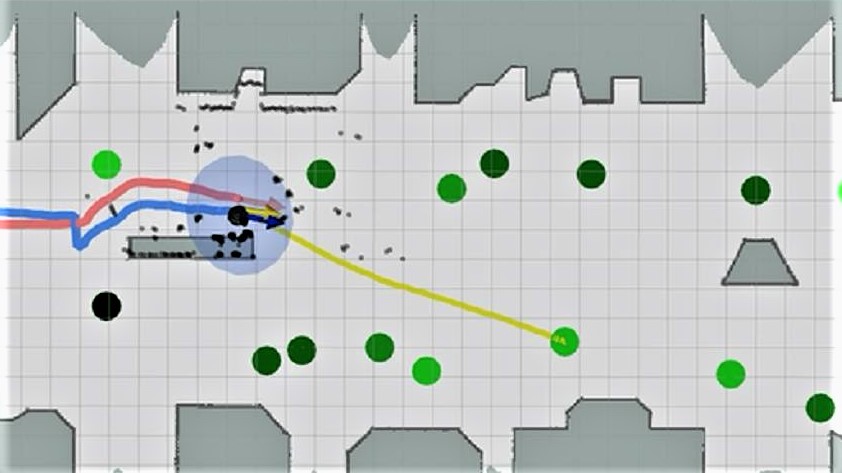}
\caption{}
\label{fig:adaptive_b}
\end{subfigure} 
\begin{subfigure}{0.23\textwidth}
\includegraphics[trim=10 0 10 0, clip, width=1.0\linewidth]{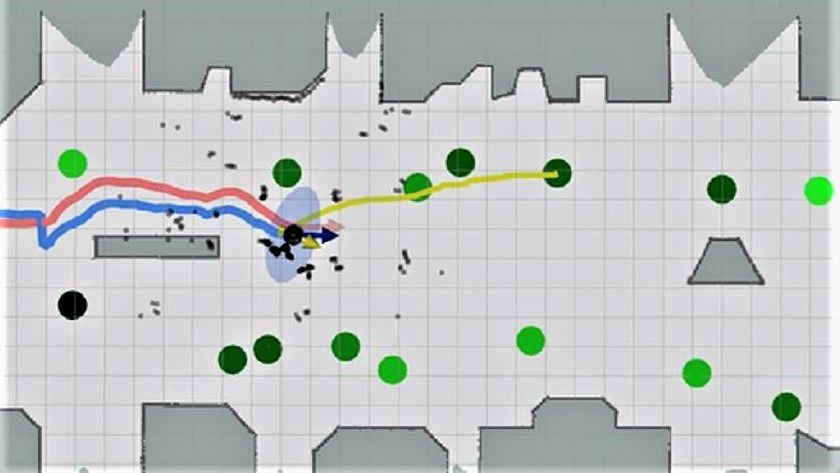}
\caption{ }
\label{fig:adaptive_c}
\end{subfigure}
\begin{subfigure}{0.23\textwidth}
\includegraphics[trim=10 0.5 10 0.5, clip, width=1.0\linewidth]{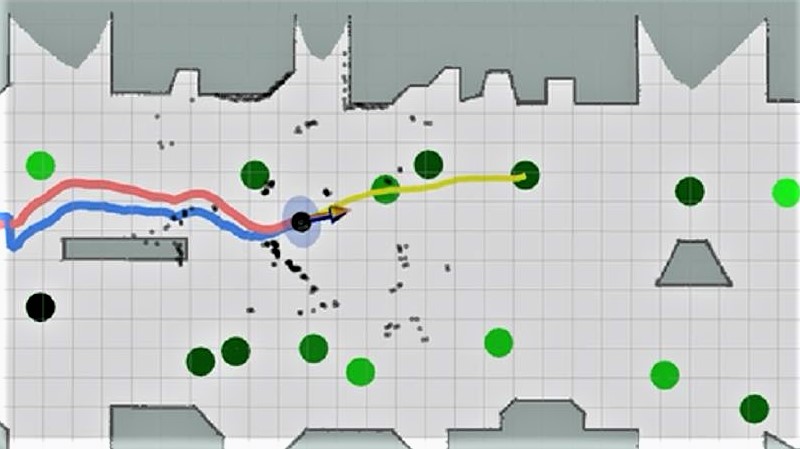}
\caption{}
\label{fig:adaptive_d}
\end{subfigure} 
\caption{Demonstration of the adaptive recovery point selection in a highly dynamic environment.}
\label{fig:adaptive_recovery_points}
\end{figure}

Note that rather than choosing one recovery point and then moving toward it until it is reached, (RL)$^2$ will determine the best recovery point in a dynamic and adaptive manner during its navigation. In particular, it may switch between recovery points according to its current observation about the pedestrian flow. For instance, in \prettyref{fig:adaptive_a} and \prettyref{fig:adaptive_b}, we can see that the bottom-right region has more free space to pass, and thus the robot will choose the recovery point in the bottom-right as the temporary goal. However, during the movement toward the goal, the bottom-right region is filling with more pedestrians, and thus the robot will adaptively switch to another recovery point that is more accessible, as shown in \prettyref{fig:adaptive_c} and \prettyref{fig:adaptive_d}. 



\begin{figure}[!htb]
\centering
\begin{subfigure}{\linewidth}
\centering
\includegraphics[trim=50 0 30 22, clip, width=\linewidth]{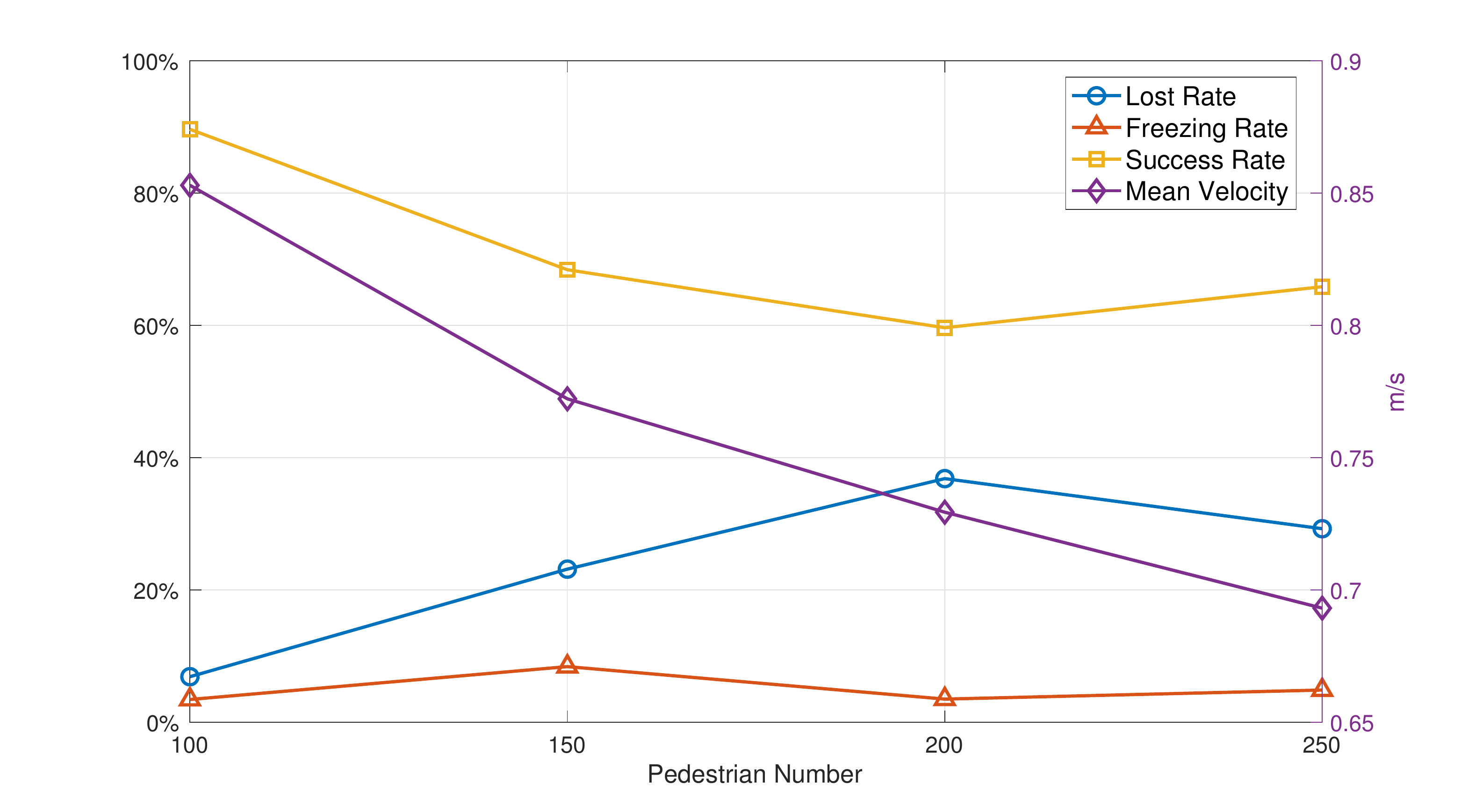}
\caption{corridor environment}
\label{fig:corridor_density}
\end{subfigure}
\begin{subfigure}{\linewidth}
\centering
\includegraphics[trim=50 0 30 22, clip, width=\linewidth]{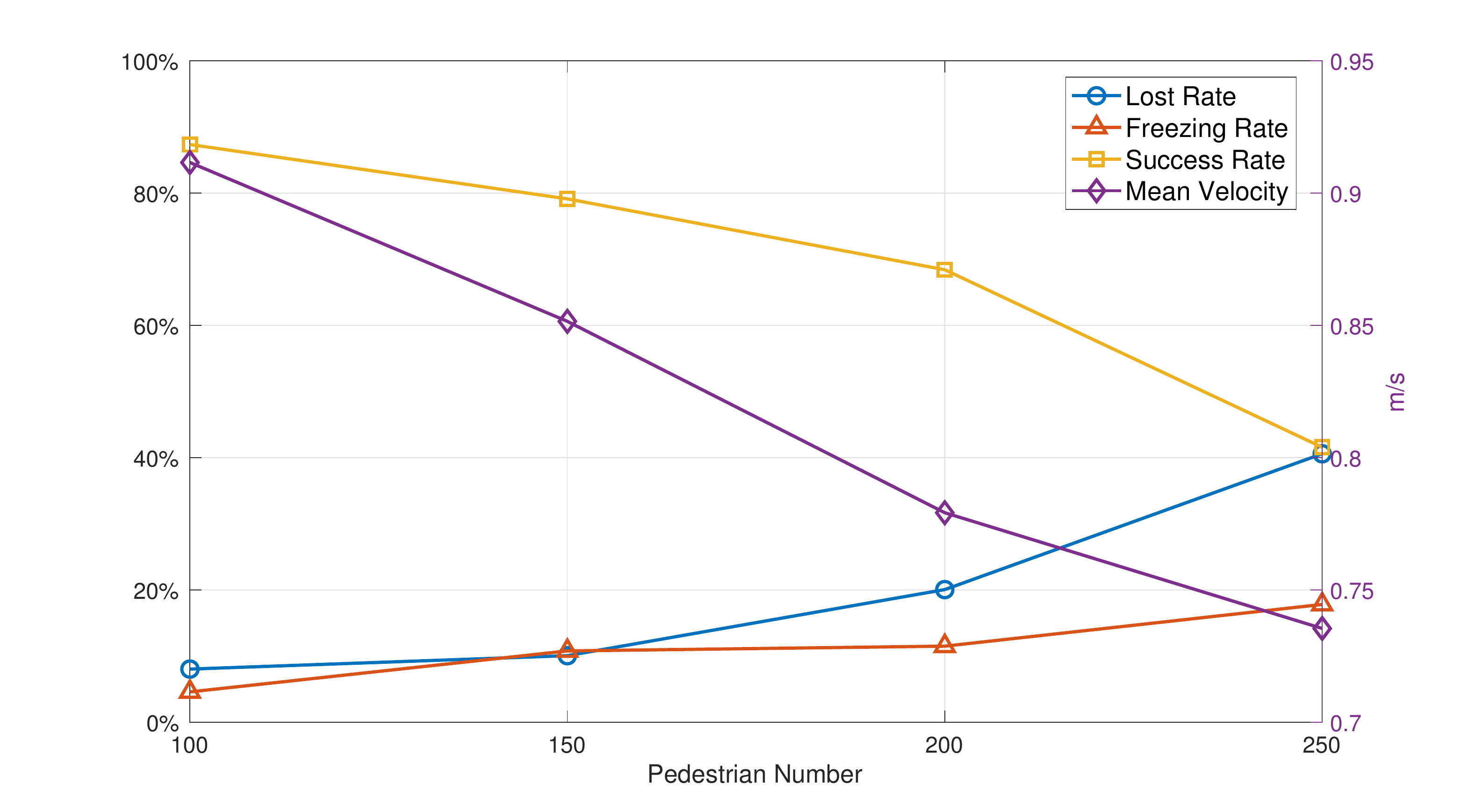}
\caption{supermarket environment}
\label{fig:supermarket_density}
\end{subfigure}
\begin{subfigure}{\linewidth}
\centering
\includegraphics[trim=50 0 30 22, clip, width=\linewidth]{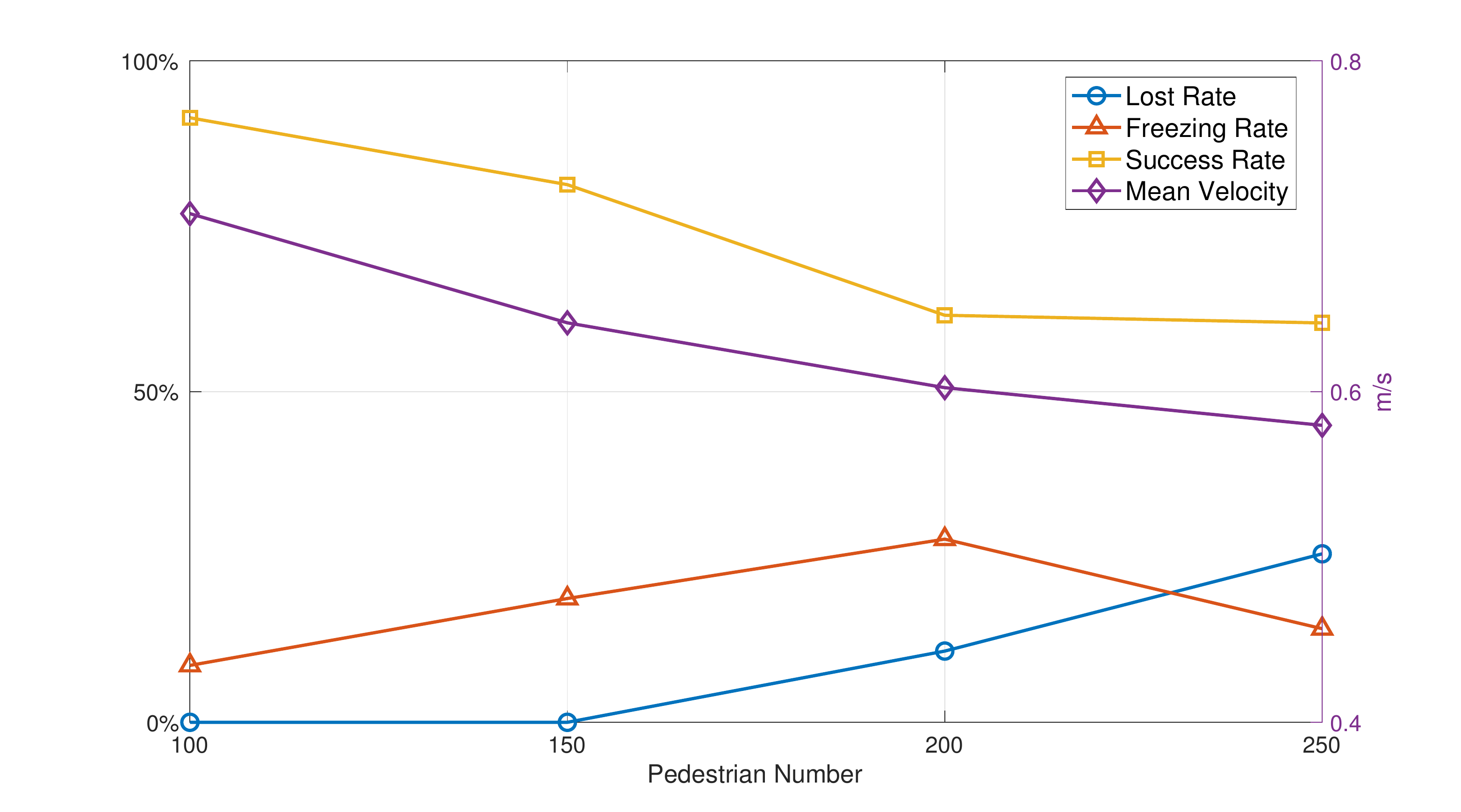}
\caption{airport environment}
\label{fig:airport_density}
\end{subfigure}
\caption{The three simulation scenarios used as the benchmark for testing navigation performance.}
\label{fig:density_exp}
\end{figure}

In \prettyref{fig:density_exp}, we show how the lost rate, the frozen rate, the success rate, and the mean velocity change when the pedestrian density in the scenario varies. In these three simulation scenarios, we gradually increase the number of pedestrians from 100 to 250. Then we can observe that, with the increase of pedestrian density, the mean velocity and the success rate of (RL)$^2$ navigation policy decrease. The frozen rate is always very small, thanks to our RL-based collision avoidance. The lost rate increases due to the increasing difficulty to have access to the recovery point. It is worth noting that narrow aisles we set in the airport environment make the robot's workspace too congested, which leads to the frozen rate higher than the lost rate .

\subsection{Real-world experiment}
\begin{figure*}[!htb]
\centering
\begin{subfigure}{0.38\linewidth}
\centering
\includegraphics[width=\linewidth]{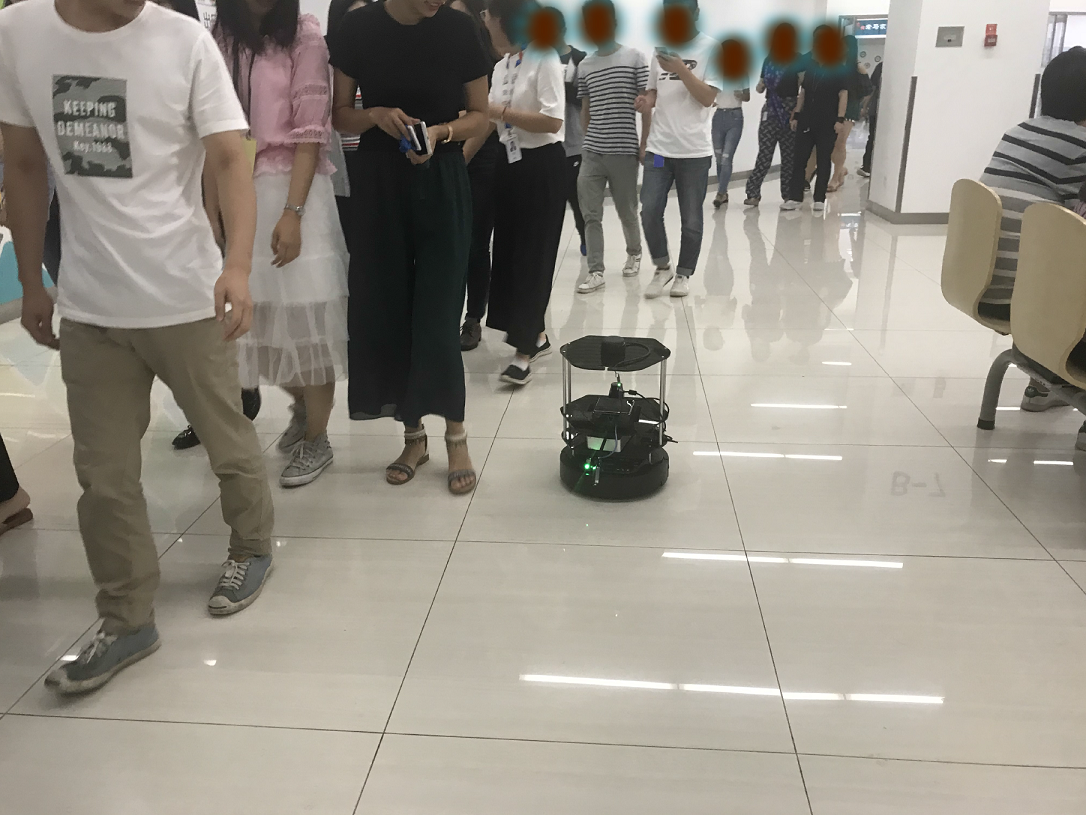}
\caption{Real scene}
\label{fig:realscene}
\end{subfigure}
\begin{subfigure}{0.60\linewidth}
\centering
\includegraphics[width=\linewidth]{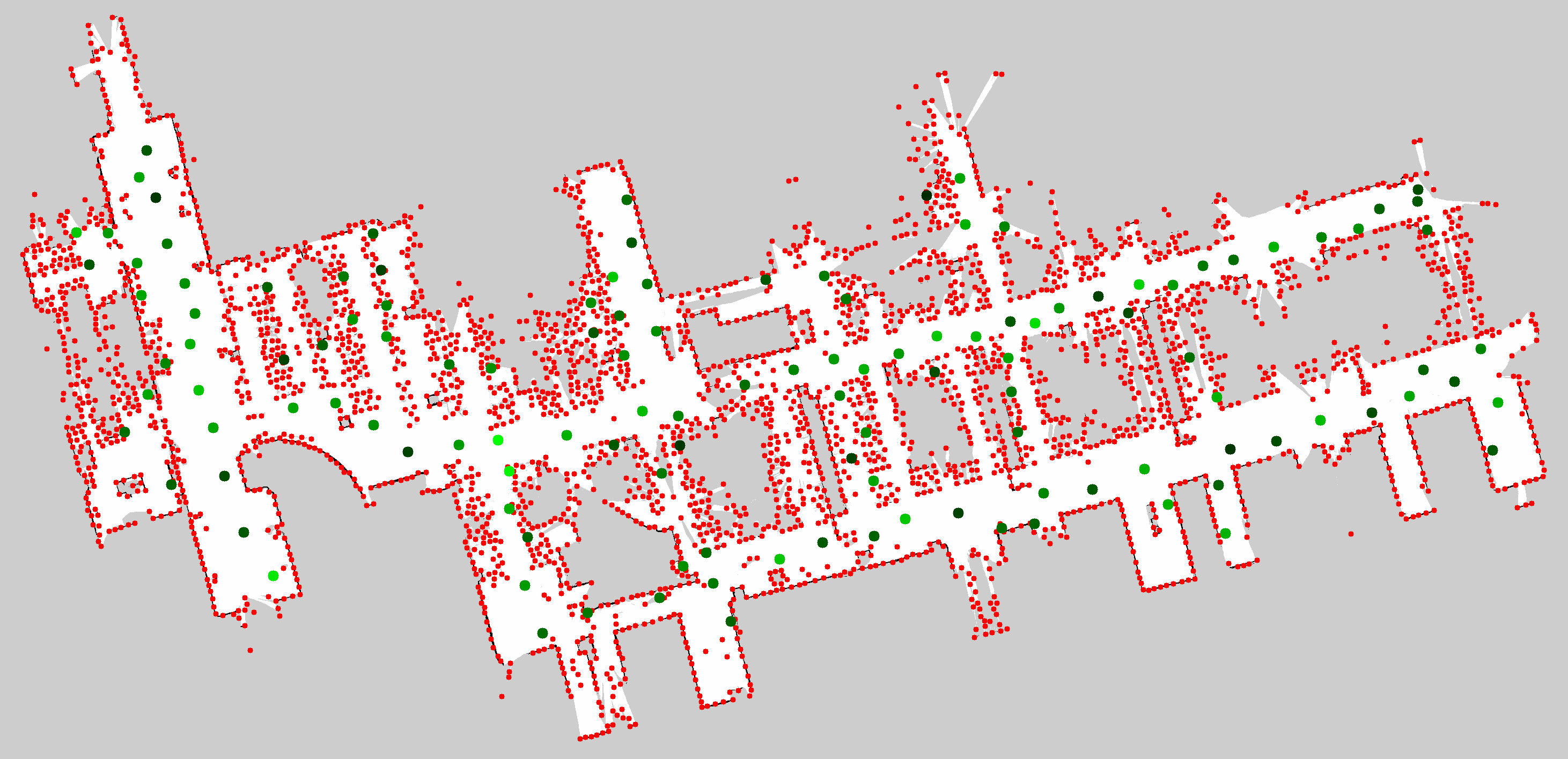}
\caption{Recovery points in the real scene}
\label{fig:realscene_recovery}
\end{subfigure}
\caption{(a) is the canteen scene for the real-robot experiment and (b) shows the map and the corresponding recovery points.}
\label{fig:real_experiment_scene}
\end{figure*}

In this part, we verify that our (RL)$^2$ framework can enable a physical mobile robot to pass through heavy crowds and arrive at the goal accurately in the real-world crowded environment. Consistent with previous simulations, we use Turtlebot2 as the robot platform and Rplidar A2 as the sensor.
We choose the canteen as our real world experimental scenario, as shown in \prettyref{fig:realscene}.
The robot is given a pre-built map as shown in \prettyref{fig:realscene_recovery} and goal position in the map. The (RL)$^2$ navigation algorithm is then executed to test whether the robot can maintain unlost and unfrozen, and can eventually arrive at the goal successfully.

\begin{figure}[!htb]
\centering
\includegraphics[width=1.0\linewidth]{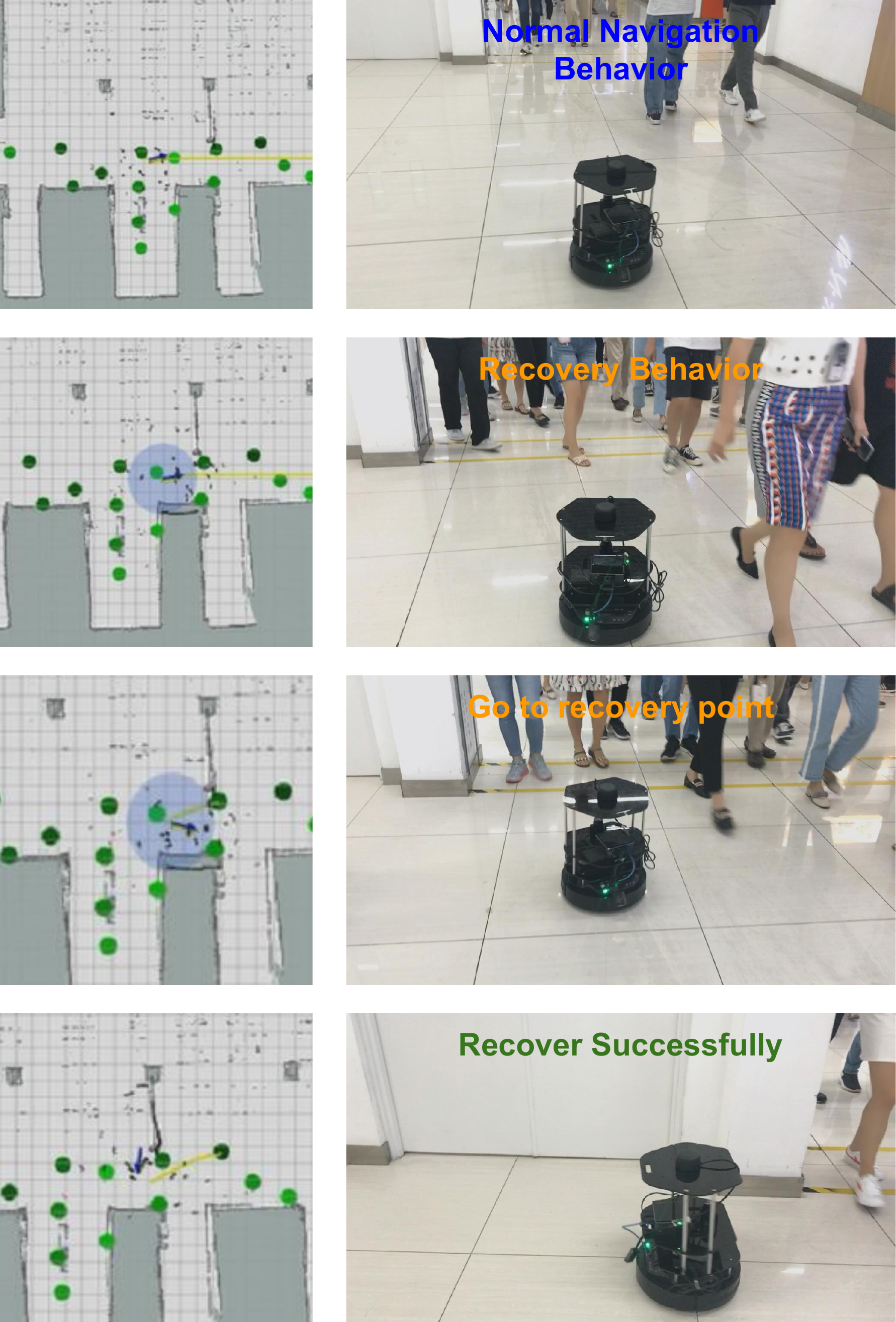}
\caption{Real-world experiment in our canteen at mealtime.}
\label{fig:real_world}
\end{figure}

The experimental results in \prettyref{fig:real_world} show that the Actor from our Actor-Critic based navigation framework can accomplish a powerful local planner responding quickly to the dynamic crowd, so that the robot does not get frozen in the crowded scene. In addition, thanks to the excellent mobility of the robot, the Critic of Actor-Critic framework is used as a high-level strategy to guide the robot toward the recovery points for active re-localization, so that the robot can achieve robust localization in the crowded environment. Please refer to \url{https://sites.google.com/view/rlslam} for more results.

\section{Conclusion}
\label{sec:conclusion}
In this paper, we propose a novel reinforcement learning based navigation framework to get a robot unlost and unfrozen in dense pedestrian crowds. In addition, we provide a benchmark including three typical scenarios with dense pedestrians. For future work, we plan to include camera resources including the depth, semantic labels, optical flows into our system to resolve our current limitation of only using 2D LiDAR information. 

{\small
\bibliographystyle{IEEEtran}
\bibliography{references}
}

\end{document}